\documentclass[compsoc]{IEEEtran}

\AtBeginDocument{%
  \providecommand\BibTeX{{%
    \normalfont B\kern-0.5em{\scshape i\kern-0.25em b}\kern-0.8em\TeX}}}

\usepackage{comment}
\usepackage{amssymb}
\usepackage{amsmath}
\usepackage[numbers]{natbib}
\usepackage{graphicx}
\usepackage{multirow}
\usepackage{xcolor}
\usepackage{pbox}
\usepackage{hyperref}

\begin{document}

\title{Neuromorphic Nearest-Neighbor Search Using Intel's Pohoiki Springs}

\author{E.~Paxon~Frady, Garrick~Orchard, David~Florey, Nabil~Imam,
Ruokun~Liu, Joyesh~Mishra, Jonathan~Tse, Andreas~Wild, Friedrich~T.~Sommer, Mike~Davies\\
{\em Intel Labs, Intel Corporation}}


\IEEEtitleabstractindextext{%
\begin{abstract}
  Neuromorphic computing applies insights from neuroscience to uncover innovations in computing technology. 
  In the brain, billions of interconnected neurons perform rapid computations at extremely low energy levels by leveraging properties that are foreign to conventional computing systems, such as 
  temporal spiking codes and finely parallelized processing units integrating both memory and computation. Here, we showcase the Pohoiki Springs neuromorphic system, a mesh of 768 interconnected Loihi chips that collectively implement 100 million spiking neurons in silicon. We demonstrate a scalable approximate $k$-nearest neighbor ($k$-NN) algorithm for searching large databases that exploits neuromorphic principles. Compared to state-of-the-art conventional CPU-based implementations, we achieve superior latency, index build time, and energy efficiency when evaluated on several standard datasets containing over 1 million high-dimensional patterns.
  Further, the system supports adding new data points to the indexed database online in $O(1)$ time unlike all but brute force conventional $k$-NN implementations.
\end{abstract}}


\maketitle

\section{Introduction}

The brain was an inspiration even for the pioneers of computing, like John von Neumann \citep{vonneumann1958}. It was  for historical and practical reasons, that the von Neumann architecture of classical computers looks very different from brains. In a traditional computer, memory (DRAM) and computing (CPU) are physically separated, information is processed according to a sequential program specification mediated by a central clock, and information is represented in digital binary strings. By contrast, in brains, information is processed fundamentally in parallel with memory and computation tightly intertwined and distributed over circuits of synapses and neurons. Although there are emergent rhythms in the brain that coordinate computation as needed, there is no central clock. Finally, communication and computation in neural circuits involve both analog and digital operations. Neurons integrate synaptic input in an analog manner, which is advantageous for efficient temporal computation, but their outputs are binary-valued spikes, which is advantageous for communication. 
Here we will demonstrate how these brain-inspired principles can be applied to perform efficient k-nearest neighbor search on Intel's Poihiki Springs neuromorphic research platform.


\section{Loihi and Pohoiki Springs}

The Loihi neuromorphic research chip is a 128-core, fully digital and asynchronous design implementing an advanced spiking neural network feature set \citep{davies2018loihi}. Its highly specialized architecture is optimized for efficiently communicating and processing event-driven spike messages. Loihi is fabricated with Intel's standard 14nm CMOS process technology. Each one of its 128 cores implements up to 1024 digital spiking neurons with features such as variable weight precision, hierarchical and compressed network routing tables, and microcode-programmed synaptic learning rules.  Additionally each Loihi chip includes three embedded x86 processors responsible for interacting with the neuromorphic cores on short timescales and converting off-chip data between conventional encodings and spikes.

Loihi also includes inter-chip communication interfaces allowing it to scale to thousands of chips in a two-dimensional mesh. Loihi has been used in several mesh-based systems to date, ranging from {\em Kapoho Bay}, a 2-chip (1x2 mesh) USB form factor device, to {\em Nahuku}, a 32-chip (4x8 mesh) custom plug-in card, to {\em Pohoiki Beach}, an early version of the Pohoiki chassis instantiating two Nahuku cards.

\begin{figure}
    \centering
    \includegraphics[width=0.9\columnwidth]{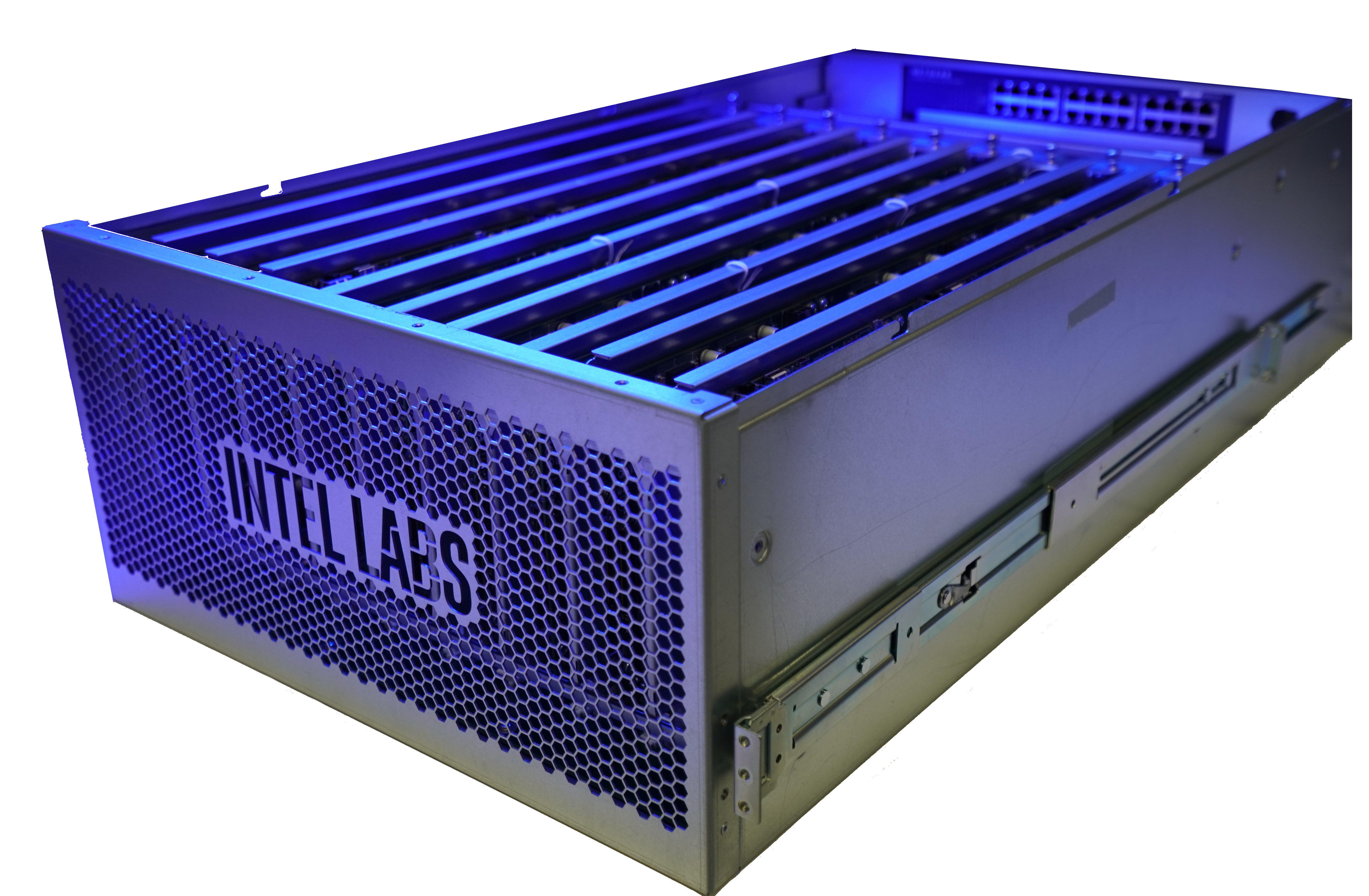}
    \caption{Pohoiki Springs 5 RU Chassis}
    \label{fig:pohoiki_chassis}
\end{figure}

Pohoiki Springs, shown in Fig. \ref{fig:pohoiki_chassis}, is the latest evolution in Loihi systems. It expands on Pohoiki Beach to a capacity of 24 Nahuku cards in a standard 19" five rack-unit chassis. 
The fully configured Pohoiki Springs chassis contains the following components:

\begin{itemize}
    \item 24 \textbf{Nahuku cards} organized into three columns of 8 cards each, for a total of 768 Loihi chips in a 12x64 mesh.
    \item Three \textbf{Arria 10 FPGAs}, one per column of Nahuku cards, that each include both the Arria 10 SX FPGAs and ARM processors that interface with the mesh of Loihi chips. The FPGA fabric converts the ARM AXI bus to the Loihi proprietary communications protocol, and the processor implements the networking stack.  Each ARM CPU serves as the {\em host} for its allocation of Nahuku cards, responsible for data I/O and CPU-coded algorithmic interaction with its mesh of Loihi chips. The hosts communicate with the remote \emph{super host} CPU over an integrated Ethernet network.
    \item One \textbf{x86-based system}, a Core i5 CPU on an ATX motherboard form factor located in the rear of the Pohoiki Springs chassis. This x86 system, referred to as the {\em super host}, is used for orchestration, configuration, and other command and control duties. It can take also part in neuromorphic computation by injecting data into and interpreting results from the 768 chip mesh via the Arria 10 ARM hosts.
    \item One embedded \textbf{Ethernet switch} that consolidates all internal Ethernet traffic into a single interface at the rear of the chassis.
\end{itemize}

Loihi implements a barrier synchronization mechanism that allows its architecture to seamlessly scale up from one core to Pohoiki Springs' heterogeneous mesh of 98,304 neuromorphic cores and 2,304 embedded x86 cores. Whether within or between chips, all cores exchange barrier messages with their neighbors that signal the completion of each algorithmic timestep. The asynchronous, blocking nature of the barrier handshakes allow timesteps to run in a variable amount of real time depending on the amount of computation and communication the mesh collectively requires on each timestep. For pure computational workloads such as nearest neighbor classification, this feature allows the system to complete computations in the minimum time possible, providing latency and power benefits.

\section{Nearest neighbor search}

As a first demonstration of a highly scalable neuromorphic algorithm on Pohoiki Springs, we apply the neuromorphic properties introduced above to nearest-neighbor search, a problem that appears in numerous applications, such as pattern recognition, computer vision, recommendation systems, and data compression. Given a database of a large number $M$ of $N$-dimensional data points, the $k$-nearest neighbor ($k$-NN) algorithm maps a specified search key to the $k$ closest matching entries. 


The performance of different $k$-NN algorithms is measured by the time complexity of a search, as well as the time and space complexity required to prepare and store the data structures used to perform the search, referred to as the search index.


There are several exact $k$-NN implementations, such as those based on space partitioning, for example using k-d trees or R trees \cite{he2012computing, cheung1998enhanced}.
However, exact approaches suffer from the curse of dimensionality \citep{Charikar2002} and for large high-dimensional databases they are too computationally expensive to use in practice on conventional hardware. 
In recent years a variety of efficient approximate $k$-NN implementations have been developed and are in wide use today. These employ diverse approaches such as dimension reduction, locality sensitive hashing, and compressed sensing \citep{Andoni2015, Har-peled2012,Charikar2002}. Recent efforts to fairly benchmark these methods have shown that even these approximate methods must choose between minimizing either query time or index preparation time \citep{aumuller2017ann}. 

Here we focus on the case of nearest neighbor search on the unit sphere where distance refers to angular distance or cosine similarity between (normalized) vectors. For this distance metric, exact nearest neighbor search can be performed by a matrix vector product (MVP) between a large data matrix of high dimensional pattern vectors and the search key.  
Given data matrix $\mathbf{D} \in {\mathbb{R}}^{N \times M}$ and search key $\tilde{d} \in {\mathbb{R}}^{N}$, the matrix vector multiplication approach to $k$-NN yields
a score vector of matches, the {\it match vector}:
\begin{equation}
\mathbf{m} = \mathbf{D}^{\top} \tilde{\mathbf{d}}   
\label{eqn:m_in_prod}
\end{equation}
The entry of the match vector with maximum amplitude identifies the nearest neighbor. For the $k$-NN problem, the set of $k$ components with largest amplitudes represents the solution. 



Our simple approximate algorithm computes and searches the matrix-vector product (\ref{eqn:m_in_prod}) on neuromorphic hardware. By encoding the data using spike-timing patterns, we can implement $k$-NN on Pohoiki Springs at large scale. 


\subsection{Data encoding}

Coding conventional data in a manner that is amenable for sparse, spike-based processing is a key aspect of neuromorphic algorithm design. We explain our approach using the Tiny Images dataset as an example. Similar processing is applied to GIST-960 \citep{Jegou2011} and GloVe \cite{pennington2014glove} datasets, and can also be applied to other datasets.
For the Tiny Images dataset, the data dimensionality is $N = 32 \times 32 \times 3 = 3072$ pixels per image, and the number of data points $M$ will scale up to $10^6$. 
To start, the data is mean-centered and normalized.

Image and other data can be reduced in dimensionality quite easily, often with minimal loss of information. In this work, we use PCA and ICA to transform input data patterns to lower dimensional representations. For large datasets, a representative subset can be used to compute the transform matrix $\mathbf{D}_{PCA} = \mathbf{U}_{PCA} \mathbf{\Sigma} \mathbf{V}_{PCA}$. A subset of $20,000$ training data points from Tiny Images were used to compute the principal components $\mathbf{V}_{PCA} \in \mathbb{R}^{N_C \times N}$, with the top $N_C = 500$ kept for dimensionality reduction, down from 3072.

\begin{figure}
    \centering
    \includegraphics[width=0.45\textwidth]{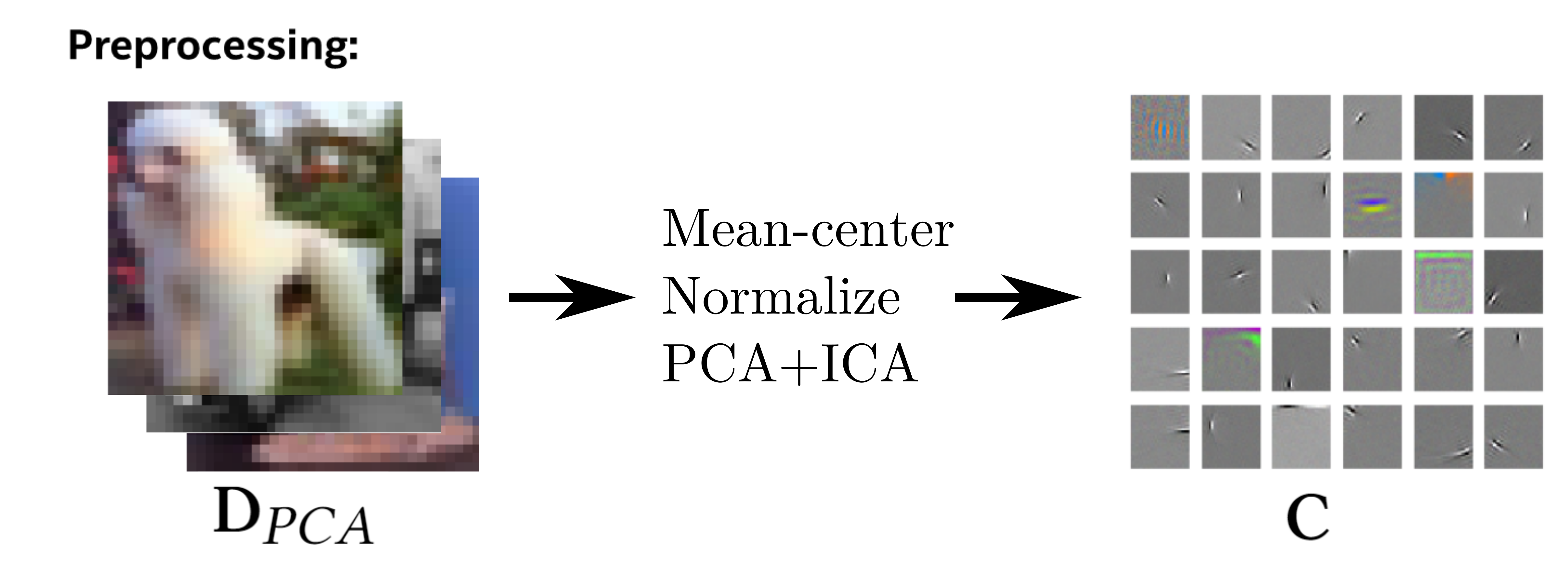}
    \caption{In the preprocessing phase, a subset of the data is used to compute the encoding matrix $\mathbf{C}$, which is used to produce a sparse representation of the data in a reduced dimensionality.}
    \label{fig:ica_compression}
\end{figure}

Following PCA reduction, the fast ICA algorithm \citep{Hyvarinen1999} is used to find the ICA mixing matrix $\mathbf{M}_{ICA} \in \mathbb{R}^{N_C \times N_C}$.
The mixing matrix is a unitary matrix that rotates the image into a basis with sparse coefficients. The PCA-ICA combination provides an encoding matrix,
\begin{equation}
  \mathbf{C} = \mathbf{M_{ICA}} \mathbf{V_{PCA}}  
  \label{c_trafo}
\end{equation}
The matrix $\mathbf{C}$ is computed offline once and stored for later online use to encode search keys $\mathbf{\tilde{d}}$. 
Specifically, an image is represented as the sparse coefficients of the ICA basis vectors (Fig.~\ref{fig:ica_compression})
\begin{equation}
    \mathbf{\tilde{e}} = \mathbf{C} \mathbf{\tilde{d}}
    \label{eqn:reduced_input}
\end{equation}

The vector $\mathbf{\tilde{e}}$ is a sparse representation of the image in a reduced dimensionality, and $k$-NN can be performed in the lower dimensional space. To do so, we encode the dataset in this reduced space, with
\begin{equation}
    \mathbf{E}=\mathbf{C} \mathbf{D}
    \label{eqn:reduced_data}
\end{equation}
Dot products in this reduced space will then remain very close to the true dot product, with $\mathbf{E}^\top \mathbf{\tilde{e}} \approx \mathbf{D}^\top \mathbf{\tilde{d}}$. Without dimensionality reduction, where $N_C=N$, these dot products would be exact. By choosing $N_C<N$, we (1) lower the computational cost of the nearest neighbor search with minimal accuracy loss and (2) obtain a sparse lower dimensional encoding of the search key that may be efficiently transferred to Pohoiki Springs as spikes.

\subsection{$k$-NN with spiking neurons}

Our neuromorphic algorithm computes $k$-NN classification with a single layer of integrate-and-fire neurons, where each neuron's membrane voltage represents the match of a particular data point with the search key. The synaptic weights feeding into a neuron encode the stored data point, and the spike timing of presynaptic spikes represent a search key. By pruning small components from the sparse search key representation $\mathbf{\tilde{e}}$, we reduce the amount of information that has to be communicated by spikes without significantly degrading the accuracy. 

To represent search keys with spike timing, we adopt previous approaches of spike time latency codes, in which earlier spikes represent larger magnitudes \citep{Hopfield1995, Thorpe1996}. 
To represent negative amplitudes, the number of inputs is double the dimension of the vector $\mathbf{\tilde{e}}$. Negative amplitudes are turned into positive amplitudes and represented as dual components in the second half of the input vector. Thus large positive and negative amplitudes in the coefficients will both result in early spikes. The inputs therefore have antagonistic receptive fields, like `on-cells' and `off-cells' seen in neuroscience. 

A search key $\mathbf{\tilde{e}} \in \mathbb{R}^{N_C}$ is represented by a spike pattern $\mathbf{\tilde{s}}(t) \in \mathbb{R}^{2N_C}$ within the input window $T$. In this demonstration the window length is $T=60$ timesteps,
\begin{equation}
    \tilde{s}_i(t) = 
    \begin{cases}
        \delta\left(t - T (1 - \tilde{e}_i/\tilde{e}_{max}) \right) & \text{if} \  \tilde{e}_i > \theta_e \\
        \delta\left(t - T ( 1 + \tilde{e}_{(i-N_C)}/\tilde{e}_{max}) \right) & \text{if} \  -\tilde{e}_{(i-N_C)} > \theta_e \\
        0 & \text{otherwise}
    \end{cases}
    \label{eq:s}
\end{equation}
where $\tilde{e}_{max} = \mbox{max } |\tilde{e}_i|$ and $\delta(t)$ denotes the Kronecker delta function over discrete timesteps $t \in [0,T]$. Note that the neurons $1,\ldots, N_C$ encode positive $\tilde{e}_i$, while the neurons $N_C +1,\ldots, 2\cdot N_C$ encode negative $\tilde{e}_i$.  
The larger the absolute value of $\tilde{e}_i$, the earlier the corresponding spike.
Pruning of small components is implemented by the threshold variable $\theta_e > 0$. Components with absolute values $|\mathbf{\tilde{e}}|/\tilde{e}_{max}$ smaller than the threshold are dropped. This reduces spike traffic on the hardware, but of course not all information of the input vector is transmitted. The used setting $\theta_e = 0.1$ typically removes about one quarter to one third of spikes.

        

The synaptic weight matrix is a concatenation of the preprocessed data matrix $\mathbf{E}$, and a sign-inversed version of it: $\mathbf{W} = [\mathbf{E}, -\mathbf{E}] \in \mathbb{R}^{2 N_C \times M_s}$.

\begin{figure}
    \centering
    \includegraphics[width=0.45\textwidth]{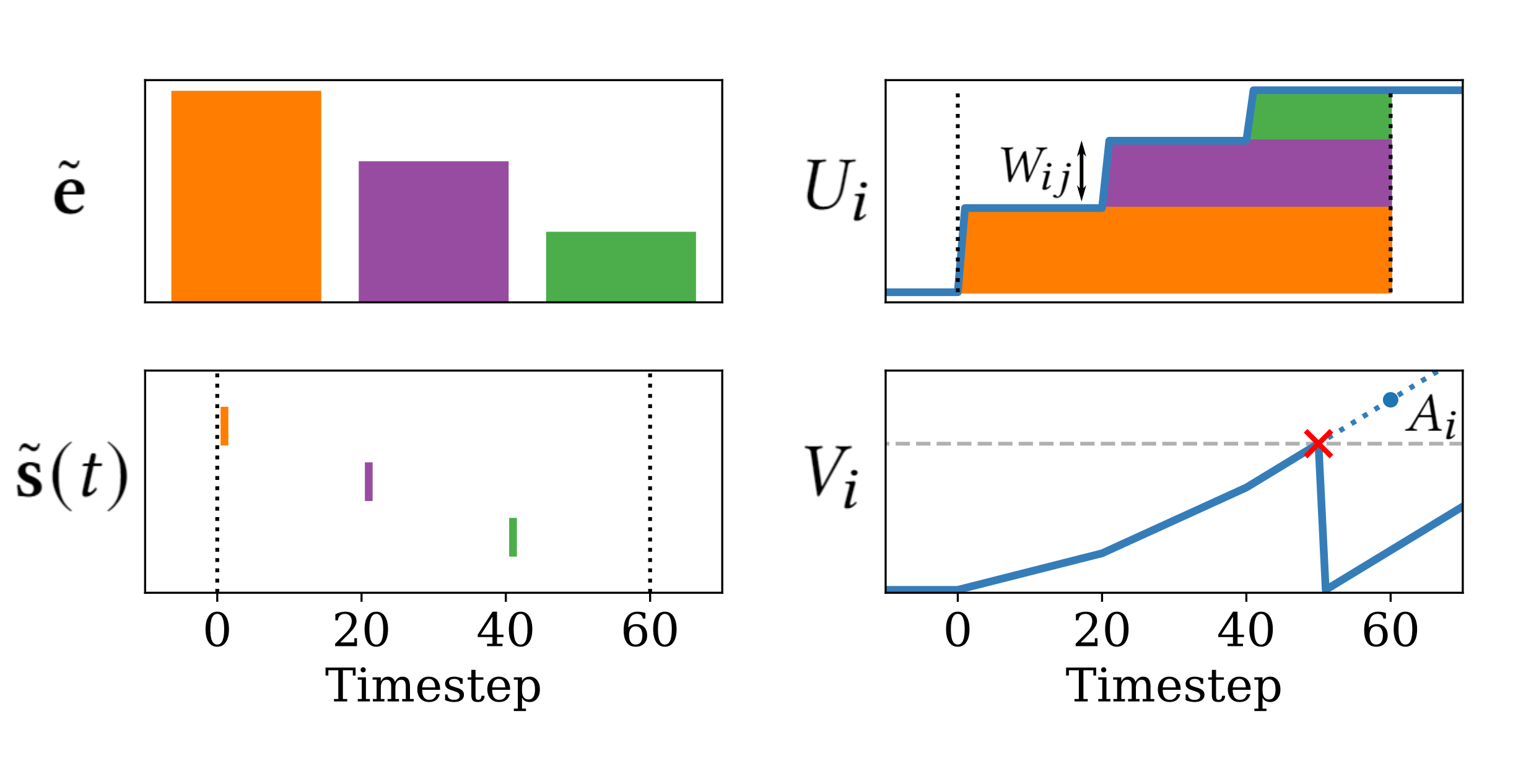}
    \caption{Computation of dot product with temporal coding.}
    \label{fig:integration}
\end{figure}

Each input spike is broadcast to all pattern match neurons where it is weighted by the synaptic strength and integrated to each neuron's synaptic current. The postsynaptic currents are again integrated in a standard integrate-and-fire neuron. To perform these computations, the Loihi chip is configured to implement neurons with the following discrete time dynamics: 
\begin{eqnarray}
        U_i (t + 1) &=& U_i(t) + \sum_j W_{ij} \tilde{s}_j(t) \label{eq:u}\\
        V_i (t + 1) &=& V_i(t) + U_i (t) \label{eq:v}
\end{eqnarray}
$U_i$ and $V_i$ represent the synaptic current and voltage in neuron $i$. When the voltage crosses threshold $\theta_V$, the neuron emits a spike and $V_i$ is reset to 0. A long refractory period prevents pattern match neurons from spiking more than once.

The spike encoding of the search key (\ref{eq:s}) together with synaptic multiplication and neuronal integrate-and-fire dynamics lead to a temporal code of the output spikes that reflect the order of the dot products between search key and the data points, $\mathbf{E}^\top \mathbf{\tilde{e}}$. Note that the area under the curve of a neuron's synaptic current, (\ref{eq:u}), can be written as $A_i = \sum_{t=0}^T \sum_j W_{ij} \tilde{s}_j = \sum_j E_{ij} \tilde{e}_j/\tilde{e}_{max}$, which is proportional to the dot product. The quantity $A_i$ is computed by the integration of the current in the voltage variable (\ref{eq:v}) (Fig. \ref{fig:integration}, right). The temporal order of output spikes, generated when the voltages exceed threshold $\theta_V$, reflects the approximate order of matches in the search.      
Thus, the detection of the first $k$ output spikes implements $k$-NN classification. 
Neurons that are too weakly activated by the search key to surpass the threshold $\theta_V$ do not spike at all. They represent weak matches that are excluded from even being ranked. 

The approach presents tradeoffs between computation precision, energy consumption, and time, which 
can be adjusted to a particular problem by parameter settings: 
\begin{itemize}
    \item Threshold $\theta_e$ governs the trade-off between sparsity in input spike patterns and representing components of the search key $\mathbf{\tilde{e}}$ with small absolute values. 
    \item Threshold $\theta_V$ governs the 
    average integration time and thereby the precision of the returned match list. Thus, raising the threshold increases precision at the cost of compute time.
    \item Length of input window $T$ determines the discretization error in the representation of the search key components represented by spike times.
    \item Synaptic resolution determines the discretization error in the representation of the stored data points.
    \item Threshold $\theta_W$ for synaptic pruning.
\end{itemize}

The adjustments of certain parameters should be coordinated for achieving best performance at minimal resource use: Integration window and synaptic resolution determine the resolutions for search key and data points, respectively. It is reasonable to choose similar resolutions for search key and data points. Similarly, input threshold and synaptic pruning threshold should be adjusted to similar cut-off levels. The dot product is computed most precisely in neurons that spike exactly at the end of the input window, so the spike threshold should be tuned jointly with the input window.

\section{Implementation}

To map nearest neighbor search to Pohoiki Springs, we take a modular approach. Subsets of the data are stored on individual Loihi chips, and the full database is distributed across the 768 chip mesh. The module defines the architecture for one chip, \emph{e.g.\/} it sets neuron model parameters, sets the weights, and instantiates the neurons used to broadcast the input keys.

To execute a search, the input query vector is converted into a temporal spike code, and a network of routing neurons distributes the query spikes throughout the mesh. The similarity comparison is computed by a layer of output neurons that integrate the contribution of the input spikes. The spike times of the output neurons are detected by spike counters in the x86 processors embedded in each Loihi chip, which send the results back to the hosts and super host over message-passing channels. Finally, the super host merges and filters the top $k$ matches based on the timestamps attached to the returned messages.

\subsection{Single chip nearest neighbor search}

The $k$-NN module for a single chip consists of $1,000$ spiking inputs, and $2,400$ spiking output neurons, with almost full connectivity. 
A subset of $M_s=2,400$ patterns will be stored on each chip in the weights between input and output neurons. The data to store is represented by the matrix $\mathbf{D}_s$, with dimension $N \times M_s$. 
This data is encoded by the matrix $\mathbf{E}_s=\mathbf{C} \mathbf{D}_s$, as in (\ref{eqn:reduced_data}). 

\begin{figure}[h]
    \centering
    \includegraphics[width=0.45\textwidth]{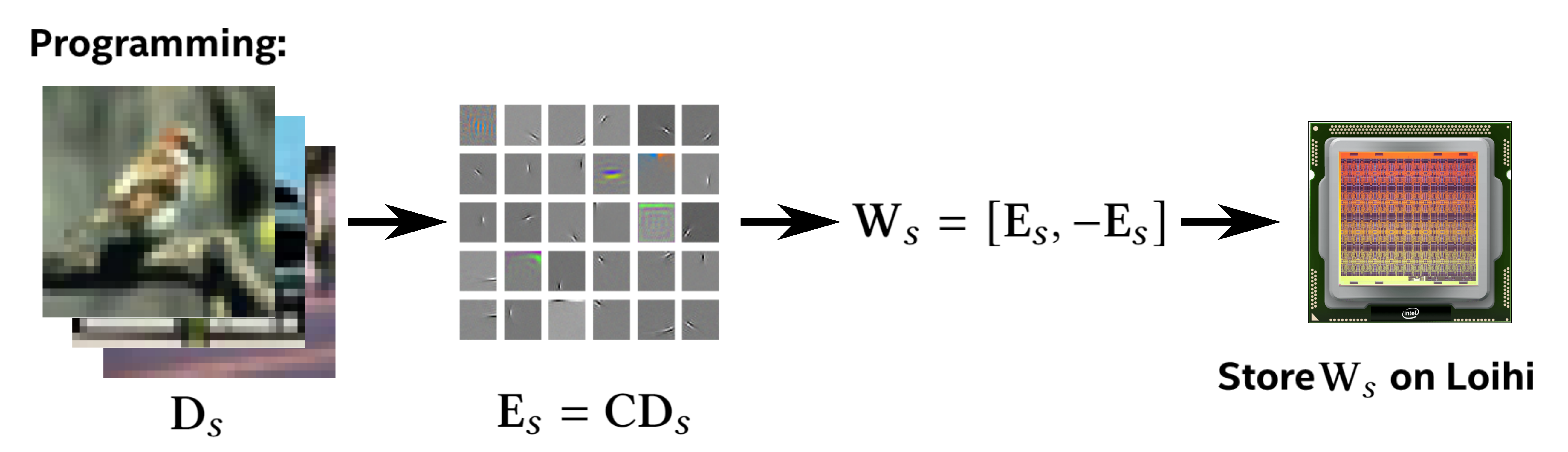}
    \caption{In the programming phase, image data is encoded and stored on individual Loihi chips. The process is repeated for each chip.}
    \label{fig:my_label}
\end{figure}

\begin{figure}[h]
    \centering
    \includegraphics[width=0.35\textwidth]{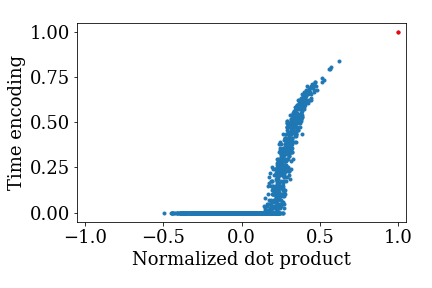}
    \caption{The normalized spike times are compared to the empirical match value.}
    \label{fig:correlation}
\end{figure}

The input weight matrix $\mathbf{W}_s=[\mathbf{E}_s, -\mathbf{E}_s]$ is encoded in the same manner as the input search coefficient vector, and is correspondingly doubled in length to match the negative components. 
The weights $\mathbf{W}_s$ are rescaled to the range $[-128, 127]$, rounded to integer values and stored as synaptic weights on a chip. We tuned the system such that the best matches would produce spikes around timestep 60.
The relationship between the timing of the output spikes from a single chip query and the dot product is visualized in Fig. \ref{fig:correlation}.



\subsection{Distributing the search over many Loihi chips}

\begin{figure*}[t]
    \centering
    \includegraphics[width=\textwidth]{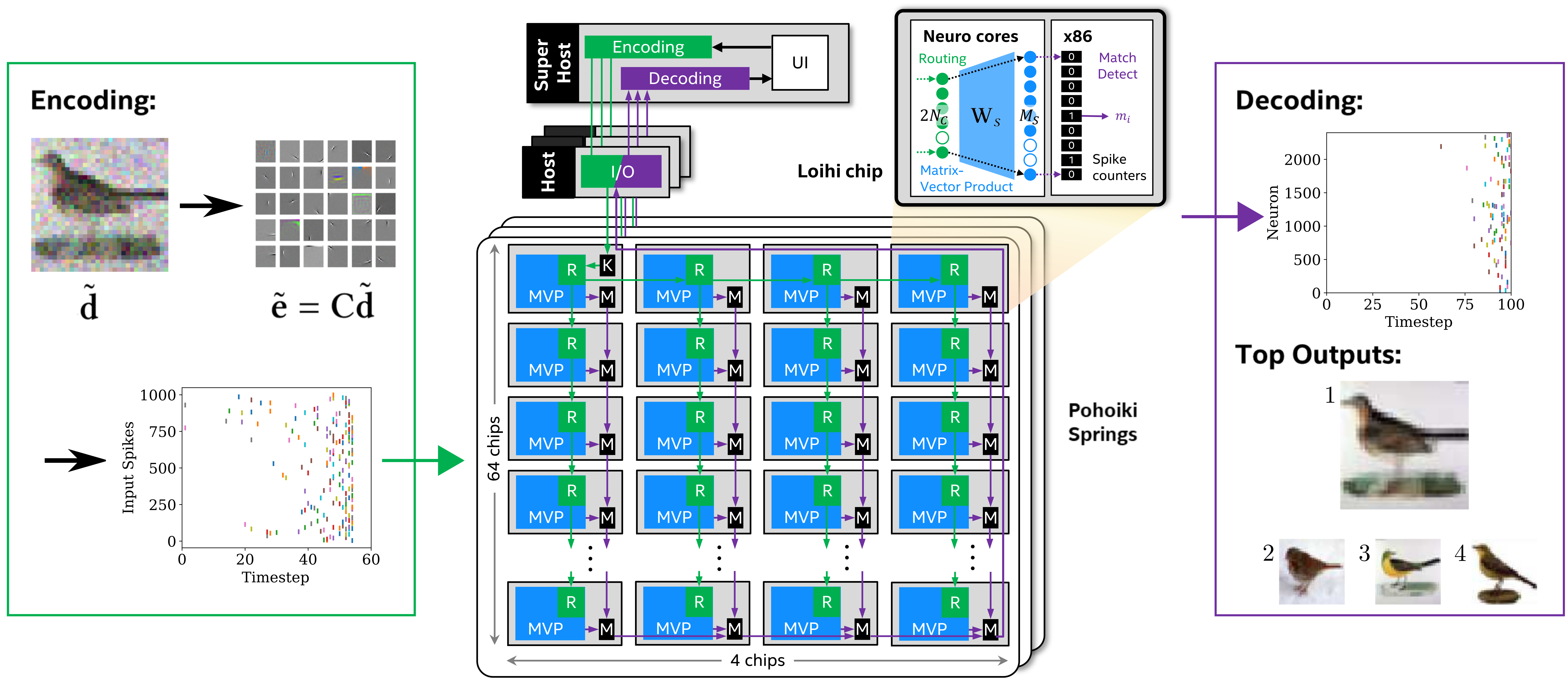}
    \caption{Nearest neighbor search on Pohoiki springs.}
    \label{fig:example}
\end{figure*}

The full pipeline of the execution phase is illustrated in Fig. \ref{fig:example}. Given a query vector $\mathbf{\tilde{d}}$, the super host CPU computes $\mathbf{\tilde{e}} = \mathbf{C}\mathbf{\tilde{d}}$ and $\mathbf{\tilde{s}}(t)$, which is then sent to the Pohoiki Springs host CPUs for further distribution into the Loihi mesh as a list of spike indices and spike times. Each host sends $\mathbf{\tilde{s}}(t)$ to the first x86 processor embedded in its column of 256 Loihi chips (Fig. \ref{fig:example}, K). 
The embedded processor K injects the spikes into a network of routing neurons (Fig. \ref{fig:example}, R) that distribute the spikes to the routing neurons in neighboring Loihi chips.  The neighboring chips in turn further route these spikes to their neighbors, propagating throughout the column as a wavefront of query spikes, advancing one layer of chips per timestep.

The routing neurons in each chip also project to a local population of integrate-and-fire neurons implementing that chip's similarity calculations over its stored patterns. The spikes activate each subset $\mathbf{W_s}$ of the matrix $\mathbf{W}$ in parallel on different chips throughout the mesh, influencing the temporal integration of all pattern match neurons appropriately as illustrated in Fig. \ref{fig:integration}.
(Fig. \ref{fig:example}, MVP).

\subsection{Detecting and aggregating match results}

As the pattern match neurons integrate to threshold, signifying close matches, they send spikes to hardware spike counters contained within their local embedded x86 processors (Fig. \ref{fig:example}, M) for match aggregation.  On each timestep, the processors detect candidate pattern matches by nonzero counter values and send these as messages to their neighboring processors in the same propagating wavefront manner as for the query distribution.

Each processor in the wavefront sequence asynchronously aggregates all of the results it receives before communicating the results onwards to the next processor. Any time a processor has sent $k$ results, it will stop sending messages and will communicate to the processors before it that they should also stop sending results. The final root processor on the last Loihi chip sends the fully aggregated result back to the host and super host as soon as it has $k$ results to send.

The output message from the last processor is an ordered list of the first $k$ matches (or more than $k$ if there are ties). The ordering directly reflects the order in which matches were found and does not require the host and super host CPUs to do any sorting. Each match also includes the timestep on which it was found, which can be used to identify and break ties for greater recall accuracy.

The super host is responsible for aggregating the final $k$ results by merging and filtering the three ordered lists of $k$ matches it receives from the Arria 10 ARM hosts, a negligible extra computation.  Although our search implementation seamlessly scales up to the entire 768-chip mesh, controlled by a single host, the gain in extra I/O bandwidth from partitioning the Loihi mesh into three host columns outweighs the cost of merging three sequences of $k$ matches to one. In fact, due to the highly asymmetric dimensions of each column of Loihi chips (4$\times$64), the barrier synchronization time per column of 14.3$\mu$s is only marginally faster than the barrier synchronization time across all 768 (12$\times$64) chips, 16.2$\mu$s.

For latency-optimized searches with coarse temporal discretization of the input window, typically the final timestep will include on the order of $k$ tied entries. For best possible recall accuracy, the super host can perform a final $k'$-NN search over the final tied entries, where $k'<k$ is the number needed to complete a full set of $k$ nearest neighbors.  Since the number of ties to search is orders of magnitude smaller than the size $M$ of the full dataset, this extra postprocessing step adds a small additional latency to the query, which is included in the results that follow.

\section{Experimental results}

For benchmarking, we follow the procedures described in \citet{aumuller2017ann}. Additionally, we measure and estimate power consumption in order to compare energy expenditures between different implementations. The ground truth is based on the normalized dot product (cosine distance) as computed on a CPU. We validate the algorithm on the Tiny Images dataset \citep{torralba200880}, as well as GIST-960 \citep{Jegou2011} and GloVe \cite{pennington2014glove}. 



\subsection{Performance evaluation}

\begin{figure}
    \centering
    \includegraphics[width=0.5\textwidth]{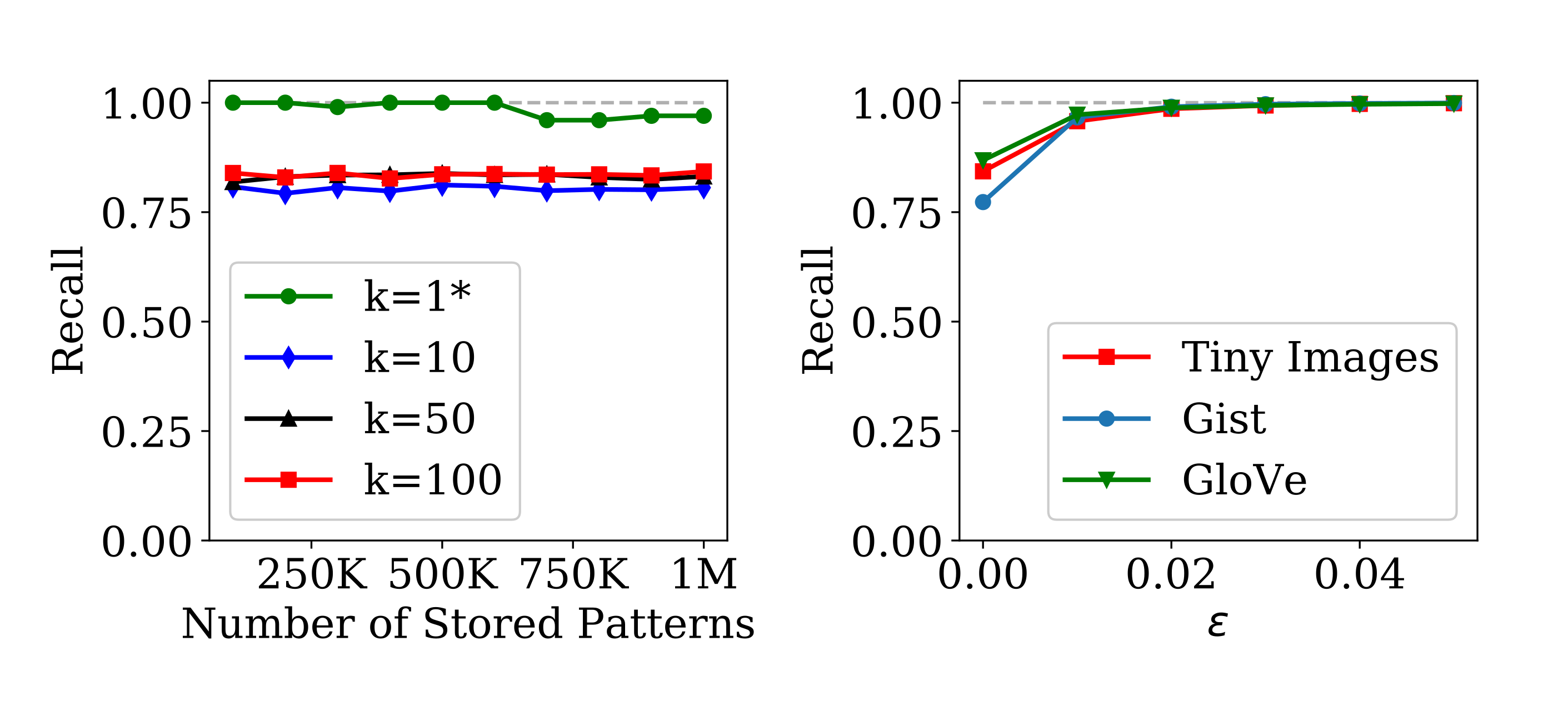}
    \caption{Recall performance.}
    \label{fig:recall}
\end{figure}

Our first experiment measures the recall performance of $k$-NN, that is, how well the algorithm returns the same results as the ground truth. In Fig. \ref{fig:recall}, left, we show the results of searching Tiny Images datasets of varying sizes from 76,800 to one million. For the $k=1$ case, we classify an input that is randomly chosen from the dataset and pixel-wise corrupted with Gaussian noise (as in Fig. \ref{fig:example}). For the other cases, we query the dataset with an input that was excluded from the dataset. Recall is calculated as the fraction of the returned $k$ data points that are no further from the search key than any data point in the ground truth top-$k$ set (Fig. \ref{fig:recall}, left).

For three one-million pattern datasets, we also evaluated the (1+$\epsilon$)-approximate recall performance, which defines an expanded window in which the top $k=100$ nearest neighbors may be found.  Fig. \ref{fig:recall}, right, shows approximate recall as a function of $\epsilon$ for Tiny Images, GIST-960 and GloVe datasets.

\begin{figure}
    \centering
    \includegraphics[width=0.45\textwidth]{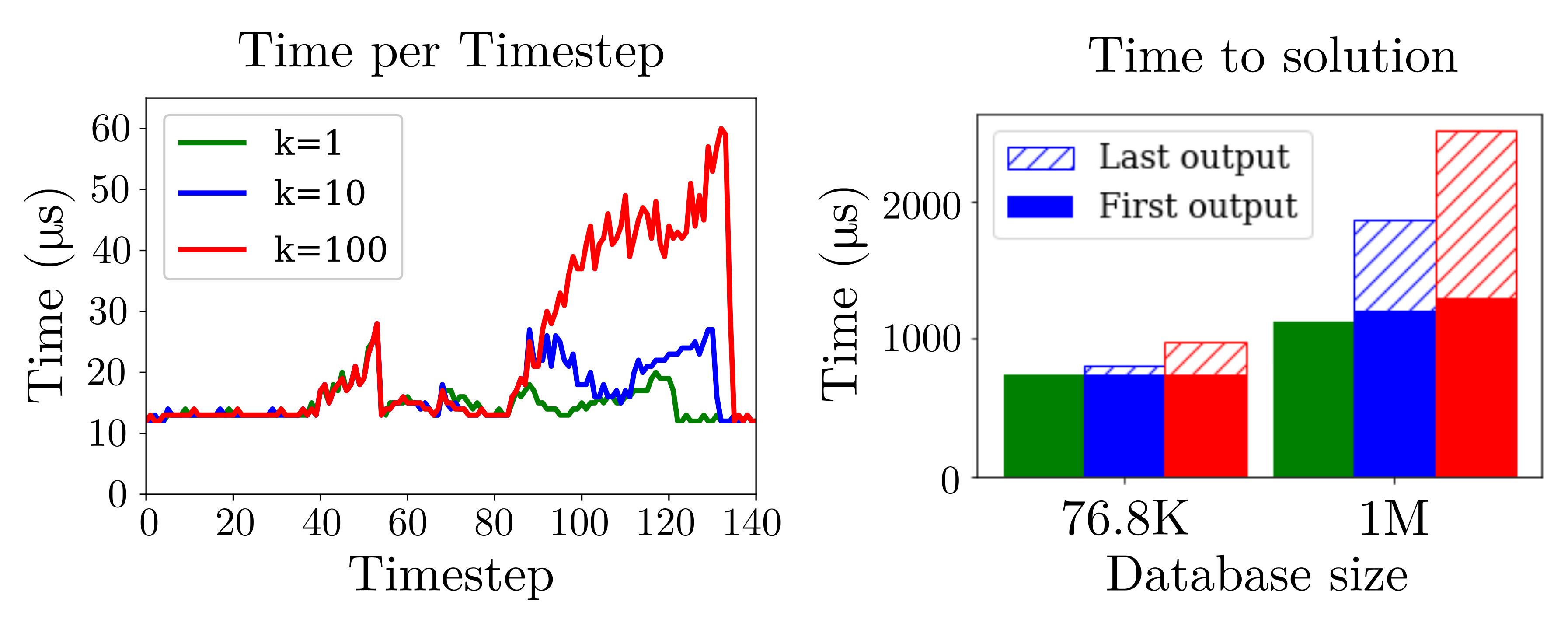}
    \caption{Search timing and latency measurements on GIST-960.}
    \label{fig:timing}
\end{figure}

We characterize the system's query latency over the range of times that Pohoiki Springs responds with the first and $k$\textsuperscript{th} match. Since the neuromorphic algorithm identifies solutions in a temporally ordered manner, the closest match is always found before the last match, and this latency spread increases with increasing dataset size (Fig. \ref{fig:timing}, right).

Depending on spike traffic, each barrier synchronized timestep of the computation can have a different duration (Fig. \ref{fig:timing}, left). In the absence of excessive spike activity, the system typically sustains just over 13$\mu$s per timestep for the 1M-pattern dataset workload and 5.8$\mu$s per timestep when processing 76,800-pattern datasets. However, slowdowns are observed during other periods. The first slowdown is noticeable near the end of the input window when many of the smaller coefficients above threshold are communicated as spikes, which have to be routed throughout the mesh. Output spikes begin to arrive near timestep 80 indicating nearest neighbors, slowing down the system. More time is needed to collect output spikes for larger $k$. Interestingly, the observed slowdowns are due to the load on Loihi's embedded x86 cores related to processing the incoming and outgoing spikes, not as a result of congestion in the neuromorphic mesh interconnect or cores.

\subsection{Power and energy}
The total power of Pohoiki Springs, including power supplies, FPGAs, ARM hosts, ATX motherboard, and Ethernet switch, is measured at the plug while running queries at maximum throughput.
Estimates of the different power components for the Loihi chips are obtained by extrapolating measurements on an instrumented board containing 32 Loihi chips and running 76,800-pattern search queries.

Table~\ref{tab:power} provides a breakdown of the Loihi mesh power consumption for a variety of sustained query workloads. \emph{Static} power is due to leakage when all circuits are fully powered. Almost all leakage can be attributed to the neuromorphic cores, which dominate chip area. The \emph{x86} power is dynamic power consumed by the x86 processors, approximately 90\% of which is idle power. \emph{Neuro} power is dynamic power attributed to the neuromorphic cores.

\begin{table}
\centering
 \caption{Power breakdown per query.}
 \begin{tabular}{|c | c c c c |} 
 \hline
 k & Size	& Static & x86 & Neuro \\
 \hline
 $1$ & 76,800 & 3.34 W & 2.09 W & 1.56 W \\
 $10$ & 76,800 & 3.34 W & 2.10 W & 1.80 W \\
 $100$ & 76,800 & 3.34 W & 2.14 W & 1.58 W \\
 $1$ & 1M & 53.4 W & 32.2 W & 16.2 W \\
 $10$ & 1M & 53.4 W & 31.7 W & 17.1 W \\
 $100$ & 1M & 53.4 W & 31.2 W & 10.8 W \\\hline
 Pohoiki wall power	& 1M & \multicolumn{3}{c |}{258 W}\\
 \hline
 CPU\textsuperscript{1} TDP	& * & \multicolumn{3}{c |}{140 W} \\
 \hline
 \end{tabular}
 \label{tab:power}
\end{table}

Table~\ref{tab:energy} further breaks down the energy consumption of a single search query. A reset phase occurs after each query to prepare the system for the next query. Total dynamic energy therefore includes both the energy required to reset and the energy required to query. The query dynamic energy is further broken down into x86 and Neuro components by isolating the embedded x86 processor workload and measuring it separately.

For extrapolation to the 1M-pattern workload, static power and x86 idle power are assumed to remain constant per chip. Neuro and x86 dynamic energy per chip (in excess of idle activity) are assumed to scale linearly with the number of Loihi timesteps. Reset energy per chip is constant for every query.

Table~\ref{tab:energy} also provides the approximate energy that a Core i9-7920X CPU\textsuperscript{1} 
requires to perform the same $\mathbf{E}^\top \mathbf{\tilde{e}}$ matrix-vector product k-NN search that Pohoiki Springs computes with spiking neurons. The energy is estimated based on the measured runtime of a NumPy float32 implementation (non-batched) multiplied by the CPU's thermal design power.

\begin{table}
\centering
  \caption{Energy breakdown per query (mJ).}
 \begin{tabular}{|c | c c c c c c | c |} 
 \hline
k & Size & Static & Reset & x86 & Neuro & Total & CPU\textsuperscript{1} \\
\hline
$1$ & 76,800 & 2.34 & 0.19 & 2.75 & 1.33 & 6.59 & \\
$10$ & 76,800 & 2.57 & 0.19 & 2.90 & 1.67 & 7.31 & \\
$100$ & 76,800 & 3.14 & 0.19 & 3.21 & 1.72 & 8.25 & \\
$1$ & 1M & 102 & 3.06 & 61.6 & 31.0 & 198 & 8,978$^\dagger$ \\
$10$ & 1M & 119 & 3.06  & 70.4 & 38.0 & 230 & 9,380$^\dagger$ \\
$100$ & 1M & 187 & 3.06 & 109 & 37.9 & 338 & 9,648$^\dagger$ \\
 \hline
 \end{tabular}
 \pbox[b]{\columnwidth}{$^\dagger$Does not include system DRAM energy.}
  \label{tab:energy}
\end{table}

\subsection{Dataset processing and programming}

Before searches may be executed, a given dataset must be processed and Pohoiki Springs must be configured. This happens over a series of three steps: (1) a dataset preprocessing step to compute the encoding matrix $\mathbf{C}$, (2) an index build step to compute the weights to be programmed into Pohoiki Springs, and (3) a programming step that writes all computed weights to the Loihi mesh. 

Dataset preprocessing entails computing PCA and ICA on a subset of the dataset. This step optimizes the data encoding for the Pohoiki Springs algorithm and only needs to be computed once per class of data. For data with a few thousand dimensions such as images, typically a subset of 10 thousand or more is needed. Here, we use 20,000 samples for computing the encoding matrix. It takes 68, 71, and 132 seconds to compute the $\mathbf{C}$ matrices for GloVe, GIST-960 and Tiny Images, respectively, using an Intel Core i9-7920X CPU\textsuperscript{1}.

The index build step involves transforming the given dataset by the encoding matrix $\mathbf{C}$, {\it i.e.\/} computing (\ref{eqn:reduced_data}), and writing the resulting weight submatrices $\mathbf{W_s}$ for each Loihi chip to disk. This is implemented as a batched NumPy computation for each chip's subset of the dataset.  For comparison to conventional $k$-NN implementations, we measure the time required to generate a single chip's weights and scale the time to the size of the dataset. 

In the final programming phase, the encoded dataset weights are loaded and written to each chip in the mesh along with all other routing tables and register values required to configure the $k$-NN application. This is a very slow step due to the current unoptimized state of the Pohoiki Springs I/O subsystem. The programming time for a 192-chip column was measured to be 893 seconds, or about 4.6 seconds per Loihi chip. Incrementally adding additional data points to the system requires on the order of 1ms to encode and program.

\subsection{Comparison to state-of-the-art}

In Table \ref{tab:comparison}, we compare the system's performance results on GIST-960 to state-of-the-art $k$-NN implementations,
\emph{Annoy}
\citep{Github:annoy}, 
\emph{Inverted file with exact post-verification} (IVF) \citep{Johnson17}, and \emph{Hierarchical Navigable Small World Graph} (HNSW) \citep{Boytsov13, Malkov16}. 
Comparison results were taken from \citet{aumuller2017ann}.
Note that our algorithm computes angular distance (cosine similarity) and uses angular distance as ground-truth, while the conventional implementations operate on Euclidean distances.
As a baseline reference point, we also compare performance results to our PCA/ICA-compressed brute force algorithm executed on an Intel i9-7920X CPU\textsuperscript{1}.

\begin{table*}
\centering
 \caption{Performance comparison}
 \begin{tabular}{|c | c c c c c c c|} 
 \hline
  & \pbox[b][1cm][c]{2cm}{$\epsilon$}
  & \pbox[b][1cm][c]{2cm}{\centering Recall} 
  & \pbox[b][1cm][c]{2cm}{\centering Query Latency (ms)} 
  & \pbox[b][1cm][c]{1.5cm}{\centering Throughput ($s^{-1}$)}
  & \pbox[b][1cm][c]{1.5cm}{\centering Index build time (s)} 
  & \pbox[b][1cm][c]{1.5cm}{\centering Index size (kB)}
  & \pbox[b][1cm][c]{2cm}{\centering Supports incremental insertions} \\
 \hline
 Annoy & 0.0  & 0.76 & 13.7 & 73.2
                     & \multirow{2}{*}{638}
                     & \multirow{2}{*}{5,176,296}
                     & \multirow{2}{*}{No} \\ 
 (CPU\textsuperscript{2})
       & 0.01 & 0.97 & 13.7 & 73.2 &  &  &   \\
 \hline
 IVF (FAISS)
       & 0.0 & 0.77 & 9.64 & 104 
                    & \multirow{2}{*}{1297} 
                    & \multirow{2}{*}{5,153,480} 
                    & \multirow{2}{*}{No} \\
 (CPU\textsuperscript{2})
       & 0.01 & 0.97 & 9.64 & 104 &  &  & \\
 \hline
 HNSW (nmslib) 
       & 0.0  & 0.78 & 1.41 & 710 
                     & \multirow{2}{*}{13,253} 
                     & \multirow{2}{*}{9,249,460}
                     & \multirow{2}{*}{No} \\
 (CPU\textsuperscript{2})
       & 0.01 & 0.98       & 1.41 & 710 & & & \\
 \hline
 PCA/ICA (CPU\textsuperscript{1}) 
       & 0.0 & 1.0         & 72 & 13.9 
                           & \multirow{2}{*}{30}
                           & \multirow{2}{*}{1,953,125} 
                           & \multirow{2}{*}{Yes} \\
 + batched (x100) & 0.0 & 1.0 & 2,254 & 44.4 &  & & \\
 \hline
  Pohoiki & 0.0 & 0.77 & 3.03 & 366 
                           & \multirow{2}{*}{30}
                           & \multirow{2}{*}{1,953,125}
                           & \multirow{2}{*}{Yes} \\
 Springs & 0.01 & 0.97 & 3.03 & 366 &  &  &  \\
 \hline
  \end{tabular}
  \centering
  \pbox[b][2.5em]{0.9\textwidth}{All Annoy, IVF, and HNSW numbers come from \citet{aumuller2017ann}, in particular \emph{GIST-960} dataset values from \url{https://github.com/erikbern/ann-benchmarks}.}
 \label{tab:comparison}
\end{table*}

The Pohoiki Springs query latency includes 220$\mu$s of preprocessing on the CPU to compute the ICA-transformed key $\mathbf{\tilde{e}}$, Eq. (\ref{eqn:reduced_input}), and 300$\mu$s of CPU postprocessing to exhaustively break ties in the final timestep.  These extra times contribute to the search latency but do not affect throughput since they can be computed concurrently with unrelated Pohoiki Springs queries. Conversely, search throughput is degraded by the reset time of 230$\mu$s on Pohoiki Springs that falls off the latency critical path.

Our results show that neuromorphic $k$-NN classification achieves comparable recall accuracy to the other algorithms, reporting 77-97\% of the true top $k$ results, with 3-4x better search latency and throughput than Annoy and IVF. 

Our algorithm is also favorable in its simplicity, which supports a fast index build time and the smallest memory footprint (Table \ref{tab:comparison}, Index size). Hence, while the highly query-optimized HNSW algorithm outperforms Pohoiki Springs in search speed by about 2x, it vastly underperforms it in index build time.
Further, because the Pohoiki Springs implementation organizes its index as a simple distributed array of data points, encoded by dense network weights, inserting a new point online during execution is an $O(1)$ operation that requires negligible time (feasibly under 1ms).

\section{Discussion}

Fundamentally, the computation of (\ref{eqn:m_in_prod}) and the subsequent top-$k$ search can be parallelized to a very fine level of granularity. This property is difficult to exploit with conventional architectures because communication and processor overhead come to dominate at high levels of concurrency. The Pohoiki Springs neuromorphic architecture supports sparse spiking representations and low-overhead barrier synchronization, and these features can be harnessed to provide a finely parallelized implementation of $k$-NN classification that is fast, scalable, and energy efficient. 





\subsection{Neuromorphic algorithm and data encoding}
Here we propose a simple neuromorphic implementation of $k$-NN classification using a layer of conventional spiking neurons, each one receiving inputs through synapses that represent a data point by the synaptic strength. The algorithmic innovation lies in how the input to these neurons is encoded and combined with the synaptic and neural dynamics to produce the desired computation with minimal spike traffic. We use a latency code in which larger amplitudes come early and spikes representing small amplitudes are suppressed. 
With this input encoding and leak-less integration, the resulting membrane voltages exactly represent the matches (dot products) at the end of the input window. However, the computation becomes approximate due to discretization and the translation of the membrane voltages into output spikes based on a fixed chosen threshold.  


Data preprocessing consists of PCA and ICA for dimensionality reduction and sparse spike encoding. The computation is relatively cheap since it only needs to be computed once on a representative sample of the data.  
The procedure is optional in cases where the data is already sparse with manageable dimensionality.

The search implementation uses brute-force parallelism to perform a simple dot product computation, compared to the complex hashing and search strategies of conventional state-of-the-art nearest neighbor search algorithms. 
Such a brute-force neuromorphic implementation achieves efficiency at scale on Pohoiki Springs by taking advantage of the architecture's fine granularity of distributed, co-located memory and computing elements in combination with rapid synchronization of temporally coded and integrated spike timing patterns. 

\subsection{Nearest-neighbor search results}

Our neuromorphic approximate $k$-NN implementation on Pohoiki Springs uniquely optimizes both index build time and search speed compared to state-of-the-art approximate nearest neighbor search implementations with equal recall accuracy.  Although batched implementations on both CPUs and GPUs can boost query throughputs to well beyond the levels evaluated here, up to 1,000-50,000 queries per second \cite{Johnson17}, the latencies of those implementations are 100x or more worse.

Additionally, our neuromorphic implementation supports adding new patterns to the search dataset in $O(1)$ complexity, on the timescale of milliseconds.  Conventionally, only brute force $k$-NN implementations can support $O(1)$ pattern insertion, which then comes at the cost of $O(M)$ search latency.  For the million-pattern datasets evaluated, this difference represents over 20x slower search speeds.  The ability to add to the search database without interrupting online operation may be highly desirable in latency-sensitive settings where the database needs to include points derived from events happening in real time.  Such applications could include algorithmic trading of financial assets, security monitoring, and anomaly detection in general.

The neuromorphic approach described also allows for simple adjustments to trade off performance in accuracy for improvements in latency. 
These adjustments can be made dynamically without requiring hours of index re-building \citep{Malkov16}.
By stretching or compressing the encoding of the input spike times (which is done on the CPU), as well as adjusting the thresholds of the output neurons, one may dynamically configure $k$-NN search with higher resolution or lower latency, as desired. However, accuracy is limited due to various sources of noise.  


One source of noise in the implementation comes from discretization error. 
The exact timing of input and output spikes are locked to discrete timesteps. With an integration window of 60 timesteps, the dynamic range of each input dimension is approximately six bits. Similarly, output spikes are discretized into time steps, giving finite resolution. Increasing time scale can improve this source of discretization noise, at the cost of longer execution time.

The more significant source of error is the temporal coding of output spikes, which is critical for efficiently identifying the top $k$ matches. In the computation, the desired dot product is exactly proportional to a pattern match neuron's membrane voltage only at the end of the input window. However, in order to search for near matches, the parameters must be tuned to permit spiking at times away from the exact end-of-window, thereby introducing inaccuracies.  In general, the thresholds should be tuned so that the pattern match neurons spike near the end of the integration window. 
Threshold re-tuning is easily executed and can be rapidly broadcast to all cores in the system.



The main shortcoming of the demonstrated implementation is that it only supports a dot product (cosine) distance metric. Many practical $k$-NN applications require Euclidean, Hamming, or other distance metrics. This limits the application space for the neuromorphic implementation in its current form.

\subsection{Conclusion}
The approximate $k$-NN classification implementation developed here exploits some, though certainly not all, of the fundamental neuromorphic properties of Pohoiki Springs. First, it exploits fine-grain hardware parallelism with fully integrated memory and computation. The computation of the closest $k$ matches is distributed over Pohoiki Springs' 100,000 cores in which the patterns themselves are stored.  Second, the algorithm uses the timing of events to encode information and to simplify computation. In this case, the multiply-accumulations of a conventional matrix-vector multiply operation are replaced by event-driven weight lookups and integration over time. 
Finally, the implementation intentionally introduces and exploits computational sparsity.  The algorithm transforms the input data representations to prefer zero components over nonzero ones, which the hardware then exploits by implicitly skipping all computation related to the zeros.

On the other hand, this example does not exploit many other important neuromorphic properties provided by Loihi. All weights and network parameters in the system are precomputed offline and remain static once loaded into the system. This leaves the plasticity features of the hardware untouched.  The computation is shallow and feed-forward, with the neuromorphic domain only responsible for computing a single matrix-vector product.  Search latency and dynamic energy remain dominated by von Neumann processing, which is not ideal.  In general, we expect greater gains as a greater proportion of the overall application falls within the neuromorphic domain, especially as recurrent feedback loops are introduced to accelerate convergence and to support pattern superposition. Such enhancements, the focus of ongoing work, promise to greatly boost the network’s storage capacity and performance.

Some aspects of the results suffer from a lack of optimization at both hardware and software levels, a consequence of the early prototype status of the Pohoiki Springs system.  Full utilization of the system resources would increase pattern capacity by at least 6x.  Programming times could be reduced by well over 10x with optimized software and I/O infrastructure. Much of the algorithmic latency and energy is dominated by the relatively trivial ancillary computation mapped to Loihi’s embedded x86 processors, which were only minimally customized for their role in neuromorphic interfacing. Loihi itself is research silicon and factors of performance improvement are feasible with design optimizations, especially relating to multi-chip scaling.

Nevertheless, the $k$-NN implementation prototyped here as the first application to run on Pohoiki Springs compares favorably to state-of-the-art solutions running on highly mature and optimized conventional computing systems.  The nearest neighbor search problem is central in a large variety of applications,
and this is just one of a wide space of algorithms supported by Loihi and Pohoiki Springs.  This suggests a promising future for neuromorphic systems as the technology is further matured and advanced to production standards.

\section{Methods}

All CPU performance measurements referenced in this work were obtained from two systems.  The systems, as annotated in the text, have the following properties: 
\begin{itemize}
    \item CPU\textsuperscript{1}: Intel Core i9-7920X CPU (12 cores, Hyper Threading enabled, 2.90GHz, 16.5 MB cache) with 128 GB RAM. OS: Ubuntu 16.04.6 LTS. Python version 3.5.2, NumPy version 1.18.2. Energy measurements were obtained using Intel SoC Watch version 2.7.0 over a duration of 120 seconds with continuously repeating workloads. 
    \item CPU\textsuperscript{2}: Used to obtained measurements referenced from \citep{aumuller2017ann}. As described in that work, evaluations were performed in Docker containers on Amazon EC2 c5.4xlarge instances equipped with Intel Xeon Platinum 8124M CPU (16 cores, 3.00 GHz, 25 MB cache) and 32GB of RAM
\end{itemize}
All software run on Pohoiki Springs used a development version of Intel's Nx SDK advanced from release 0.9.5.rc1.

With the exception of CPU\textsuperscript{2} measurements quoted from \citep{aumuller2017ann}, all performance results are based on testing as of March 2020 and may not reflect all publicly available security updates. No product can be absolutely secure.



\bibliographystyle{IEEEtranN}
\bibliography{references,Fritz_hdspike,knn}
\end{document}